\documentclass{article}

\usepackage[final]{nips_2017}

\usepackage[utf8]{inputenc} % allow utf-8 input
\usepackage[T1]{fontenc}    % use 8-bit T1 fonts
\usepackage{hyperref}       % hyperlinks
\usepackage{url}            % simple URL typesetting
\usepackage{booktabs}       % professional-quality tables
\usepackage{amsfonts}       % blackboard math symbols
\usepackage{nicefrac}       % compact symbols for 1/2, etc.
\usepackage{microtype}      % microtypography
\usepackage{amsmath} % assumes amsmath package installed
\usepackage{bm}
\usepackage{ dsfont }
\usepackage{footnote}
\makesavenoteenv{tabular}
\makesavenoteenv{table}
\usepackage{caption}
\usepackage{subcaption}
\usepackage{graphicx}
\usepackage{tabularx}

\title{Deep Gaussian Processes with \\ Decoupled Inducing Inputs}

\author{
  Marton Havasi \\
  Department of Engineering\\
  University of Cambridge \\
  \And
  Jos\'e Miguel Hern\'andez-Lobato \\
  Department of Engineering \\
  University of Cambridge \\
  \And 
  Juan Jos\'e Murillo-Fuentes \\
  University of Seville \\
}

\begin{document}

\maketitle

\begin{abstract}
Deep Gaussian Processes (DGP) are hierarchical generalizations of Gaussian Processes (GP) that have proven to work effectively on a multiple supervised regression tasks. They combine the well calibrated uncertainty estimates of GPs with the great flexibility of multilayer models. In DGPs, given the inputs, the outputs of the layers are Gaussian distributions parameterized by their means and covariances. These layers are realized as Sparse GPs where the training data is approximated using a small set of pseudo points. In this work, we show that the computational cost of DGPs can be reduced with no loss in performance by using a separate, smaller set of pseudo points when calculating the layerwise variance while using a larger set of pseudo points when calculating the layerwise mean. This enabled us to train larger models that have lower cost and better predictive performance.
\end{abstract}

\section{Introduction}

In this work, we are considering doubly stochastic DGPs \citep{deepgpdoubly} \citep{deepgp}. In this setting, each layer is a Sparse GP \citep{sparsegp} where the inputs are sampled from the output distribution of the previous layer (the input data in the case of the initial layer). These layers are parameterized by their covariance function and a set of inducing input-output pairs (previously referred as pseudo points). 

Generally, the inducing inputs are treated as hyperparameters and the inducing outputs (the latent variables) are variationally approximated by a multivariate Gaussian distribution. In a recent interpretation, \citep{decgp} showed the duality between GPs and Gaussian Measures. As a result, they showed that the mean and the variance computations are not required to share the inducing inputs. Since calculating the variance is significantly more costly than calculating the mean, one can speed up the training process by using a second, reduced set of inducing inputs when computing the variance. This is referred to as a Decoupled GP.

Our contribution is showing that the decoupled approach is applicable in the case of DGPs. While changes necessary to the original parameterization proposed by \citet{decgp}, the resulting DGPs were both faster and had higher predictive performance than their non-decoupled counterparts.

\section{Methods}

This section gives a brief overview of Variational Inference for Deep Gaussian Processes and explains how they can be incorporated with decoupled inducing inputs.

\subsection{Variational Inference for Deep Gaussian Processes}

Consider a set of input output pairs $\bm X$, $\bm Y$. Lets denote the layer outputs as $\bm F$ and the inducing input-output pairs with $\bm U$ and $\bm Z$ respectively. In a single layer sparse Gaussian Process (GP) \citep{sparsegp} \citep{orig_sparsegp}, one can express the joint probability density as
\begin{equation}
p(\bm Y, \bm F, \bm U)=p(\bm Y| \bm F)p(\bm F|\bm U)p(\bm U)\,.
\end{equation}

The first term, $p(\bm Y|\bm F)$ refers to the likelihood. In our case, Gaussian likelihood is used exclusively. The second part, $p(\bm F| \bm U)p(\bm U)$ is the sparse GP prior. $p(\bm U)=\mathcal{N}(\bm U|\bm 0, k(\bm Z, \bm Z))$ where $k$ is the covariance function. We will be using the notation $k(A, B)=\bm K_{AB}$ onwards.

 $p(\bm F|\bm U)$ is given by the GP model \citep{gp}:
\begin{equation}
\begin{aligned}
p(\bm F|\bm U)=&\mathcal{N}(\bm F| \mu, \Sigma) \\
 \mu=&\bm K_{\bm X \bm Z}\bm K_{\bm Z \bm Z}^{-1}\bm U \\
 \Sigma=&\bm K_{\bm X \bm X} - K_{\bm X \bm Z}K_{\bm Z \bm Z}^{-1}K_{\bm X \bm Z}^T\,.
\end{aligned}
\end{equation}

When calculating the posterior of $\bm U$ given the data $(\bm X, \bm Y)$, we are forced to approximate $p(\bm U|\bm X, \bm Y)$ by $q(\bm U)=\mathcal{N}(\bm U|\bm m, \bm S)$ \citep{sparsegp} due to tractability problems. The advantage of this form is that $\bm U$ can be marginalized and the variational posterior is simply
\begin{equation}
\begin{aligned}
q(F|\bm m,\bm S)=& \int p(\bm F|\bm U)q(\bm U)d\bm U=\mathcal{N}(\bm F| \tilde{\mu}, \tilde{\Sigma})\\
\tilde{\mu}=& K_{\bm X \bm Z}K_{\bm Z \bm Z}^{-1}\bm m\\
\tilde{\Sigma}=& K_{\bm X \bm X} - K_{\bm X \bm Z}K_{\bm Z \bm Z}^{-1}(K_{\bm Z \bm Z} - \bm S)K_{\bm Z \bm Z}^{-1}K_{\bm X \bm Z}^T\,.
\end{aligned}
\label{gpeq}
\end{equation}

Finally, according to the variational inference principles, the target of the optimization process is the Evidence Lower Bound (ELBO):
\begin{equation}
\mathcal{L}=\mathds{E}_q[\log(\bm Y|\bm F)]-KL[q(\bm U)||p(\bm U)]\,.
\end{equation}

In the multilayer setting, we employ similar approximations:
\begin{equation}
\begin{aligned}
p(\bm Y, \{\bm F, \bm U\}_{l=1}^L)=& p(\bm Y| \bm F^L)\prod_{l=1}^L p(\bm F^l|\bm U^l)p(\bm U^l)\\
q(\{\bm F\}_{l=1}^L|\{\bm m_l, \bm S_l\}_{l=1}^L) =& \prod_{l=1}^L q(\bm F^l|\bm m_l, \bm S_l)  =\prod_{l=1}^L \mathcal{N}(\bm F^l|\tilde{\mu}^l, \tilde{\Sigma}^l)
\end{aligned}
\end{equation}

where $\bf F^l$ is the output of the $l$-th layer (and input of the $l+1$-st) for $l=1 \dots L$. The input to the first layer is simply $\bm F^0 = \bm X$. The ELBO is analogously
\begin{equation}
\begin{aligned}
\mathcal{L}_{DGP}=& \mathds{E}_q[\log p(\bm Y|\bm F)]-\sum_{l=1}^L KL[q_l||p_l]\\
KL[q_l||p_l]= &\frac{1}{2}\bm m_l^{T}\bm K_{\bm Z \bm Z}^{-1}\bm m_l - \frac{1}{2}\log \frac{|\bm K_{\bm Z \bm Z}|}{|\bm S|} + \frac{1}{2}tr(\bm K_{\bm Z \bm Z}^{-1}\bm S) - \frac{M}{2} 
\end{aligned}
\end{equation}
where $M$ is the number of inducing points ($M=|Z|$).

\subsection{Decoupled Inducing Inputs}

Using the dual formulation of a GP as a Gaussian Measure, \citep{decgp} have shown that it does not necessarily have to be the case that $\tilde{\mu}$ and $\tilde{\Sigma}$ are parameterized by the same set of inducing inputs ($\bm Z$).

Instead of eq. \ref{gpeq}, a new parameterization is used that utilizes two different sets of inducing inputs $\bm Z_a$ and $\bm Z_b$. $\tilde{\mu}$ is parameterized by $\bm a=K_{\bm Z_a \bm Z_a}^{-1}\bm m$, which is beneficial, since it does not require computing the inverse covariance matrix ($K_{\bm Z_a \bm Z_a}^{-1}$). $\tilde{\Sigma}$ is defined using $(\bm B^{-1} + \bm K_{\bm Z_b \bm Z_b})^{-1}=K_{\bm Z_b \bm Z_b}^{-1}(K_{\bm Z_b \bm Z_b} - \bm S)K_{\bm Z_b \bm Z_b}^{-1}$:
\begin{equation}
\begin{aligned}
& \tilde{\mu} = \bm K_{\bm X \bm Z_a}\bm a \\
& \tilde{\Sigma} = \bm K_{\bm X \bm X} - \bm K_{\bm X \bm Z_b}(\bm B^{-1} + \bm K_{\bm Z_b \bm Z_b})^{-1}\bm K_{\bm X \bm Z_b}^T\,.
\end{aligned}
\label{decgpeq}
\end{equation}

The corresponding to the new parameterization, the ELBO takes a slightly different form:
\begin{equation}
\begin{aligned}
\mathcal{L}_{DGP}= &\mathds{E}_q[\log p(\bm Y|\bm F)]-\sum_{l=1}^L KL[q_l||p_l] \\
KL[q_l||p_l]= &\frac{1}{2}\bm a_l^{T}\bm K_{\bm Z_a \bm Z_a}\bm a_l + \frac{1}{2}\log |\bm I + \bm K_{\bm Z_b \bm Z_b}\bm B_l| - \frac{1}{2}tr(\bm K_{\bm Z_b \bm Z_b}(\bm B_l^{-1} +\bm  K_{\bm Z_b \bm Z_b})^{-1})
\end{aligned}
\end{equation}

where the KL divergence is given up-to a constant.

The reformulation greatly reduces the computational complexity. The time complexity of calculating the mean in a single layer is $O(M^3 + NM)$, the variance is $O(M^3 + NM^2)$ and the KL divergence is $O(M^3 + NM^2)$ (where $M$ is $|\bm Z|$ and $N$ is the size of the minibatch). However, after the decoupling, the cost the mean becomes $O(NM_a)$, the variance becomes $O(NM_b^2 + M_b^3)$ and the KL divergence becomes $O(NM_a + NM_b^2 + M_b^3)$. This is due to the new parameterization not requiring the costly inversion of the covariance matrix. The overall cost reduces from $O\big(L(DNM^2 + M^3)\big)$ to $O\big(L(DNM_a + DNM_b^2 + M_b^3)\big)$ (where $L$ is the number of layers and $D$ is the width of the hidden layers). Note that the cost of inverting the covariance matrix does not scale with the layer width, since every node in the same layer shares the convariance matrix. This leads to considerable improvement in training time if $M_a \gg M_b$.

\subsection{Alternative parameterizations}

Unfortunately, the parameterization advocated by \citep{decgp} (eq. \ref{decgpeq}, referred as $\tilde{\mu}_{CB}$ and $\tilde{\Sigma}_{CB}$ onwards) has poor convergence properties. The dependencies of the values of $a_l$ in the ELBO result in a highly non-convex optimization domain, which then leads to high variance gradients. This impedes convergence even at small learning rates. 

To combat the issue, we look at different parameterizations to lift the dependencies and achieve stable convergence. We consider the natural extension of the standard sparse GP parameterization to decoupled inducing inputs. For the mean, we have two alternatives. The first option is the standard sparse GP (defined in eq. \ref{gpeq}, referred as $\tilde{\mu}_{GP}$) while the second option is to precondition $\bm m$ with $\bm L^{-1}$ (shown in Table \ref{mean}, referred as $\tilde{\mu}_{GPcent}$), the Cholesky-decomposition of $\bm K$. The goal of the preconditioner is that it transforms the prior distribution to be a standard Gaussian. This aids convergence, because the term for the mean in the ELBO ($\frac{1}{2}\bm m_l^{T}\bm K_{\bm Z \bm Z}^{-1}\bm m_l$) becomes $\frac{1}{2}\bm m_l'^{T}\bm m_l'$. As for the variance, the sparse GP parameterization offers only a single option (eq. \ref{gpeq}, referred as $\tilde{\Sigma}_{GP}$) which is shown in Table \ref{var}.

As discussed earlier, $\tilde{\mu}_{CB}$ shows unstable convergence. While the problem was fixed with $\tilde{\mu}_{GP}$ and $\tilde{\mu}_{GPcent}$, this came at a cost. Both of the these require computing the inverse covariance matrix for the mean ($\bm K_{\bm Z_a \bm Z_a}^{-1}$), which leads to an increased overall cost of $O\big(L(DNM_a  + M_a^3 + DNM_b^2 + M_b^3)\big)$. Fortunately, this is still an improvement over the original cost of DGPs when $M_a \gg M_b$. Note that the cost reduction is more significant when $N$ and $D$ are large, due to the term $DNM_b^2$ dominating the overall cost. In our experiments, we used $\tilde{\mu}_{GPcent}$ although the difference in terms of performance between $\tilde{\mu}_{GP}$ and $\tilde{\mu}_{GPcent}$ was limited.

For the variance, both $\tilde{\Sigma}_{CB}$ and $\tilde{\Sigma}_{GP}$ show stable convergence and they exhibit similar performance. In our experiments, we used $\tilde{\Sigma}_{GP}$.

\begin{table}[h]
\begin{tabular}{lll}
       \toprule
 $\tilde{\mu}$ & Complexity & Convergence \\ 
 \midrule
 $\tilde{\mu}_{CB} = \bm K_{\bm X \bm Z_a}diag(\bm K_{\bm Z_a \bm Z_a})\bm a$ & $O\big(L(DNM_a)\big)$ & unstable \\ 
 $\tilde{\mu}_{GP} = \bm K_{\bm X \bm Z_a}\bm K_{\bm Z_a \bm Z_a}^{-1}\bm m$ & $O\big(L(DNM_a + M_a^3)\big)$ & stable \\ 
 $\tilde{\mu}_{GPcent} = \bm K_{\bm X \bm Z_a}(\bm L^T)^{-1}\bm m'$ & $O\big(L(DNM_a + M_a^3)\big)$ & stable \\ 
       \bottomrule
\end{tabular}
\centering
\caption{Different parameterizations of the mean. $\bm L$ refers to the Cholesky-decomposition: $\bm L\bm L^T = \bm K_{\bm Z_a \bm Z_a}$}
\label{mean}
\end{table}

\begin{table}[h]

\resizebox{\textwidth}{!}{\begin{tabular}{lll}
       \toprule
 $\tilde{\Sigma}$ & Complexity & Convergence \\ 
 \midrule
$\tilde{\Sigma}_{CB}=\bm K_{\bm X \bm X} - \bm K_{\bm X \bm Z_b}\bm K_{\bm Z_a \bm Z_a}^{-1}(\bm K_{\bm Z_a \bm Z_a} - \bm S)\bm K_{\bm Z_a \bm Z_a}^{-1}\bm K_{\bm X \bm Z_b}^T$ & $O\big(L(DNM_b^2 + M_b^3)\big)$ & stable \\ 
$\tilde{\Sigma}_{GP}=\bm K_{\bm X \bm X} - \bm K_{\bm X \bm Z_b}(\bm B^{-1} + \bm K_{\bm Z_b \bm Z_b})^{-1}\bm K_{\bm X \bm Z_b}^T$& $O\big(L(DNM_b^2 + M_b^3)\big)$ & stable \\ 
       \bottomrule
\end{tabular}}
\centering
\caption{Different parameterizations of the covariance. $\bm S$ is parameterized as $\bm S=\bm L_s \bm L_s^T$ and $\bm B$ is parameterized as $\bm B=\bm L_b \bm L_b^T$ for computational stability.}
\label{var}
\end{table}

\section{Experiments}

The goal of the experiments is to show that the decoupled approach is not only more cost efficient, but also has better performance than the original, non-decoupled approach.

The experiments are conducted on three datasets. Two benchmark UCI regression datasets: \textit{kin8nm} and \textit{protein} as well as a third regression dataset, \textit{molecules}, that has binary features and it describes the energy conversion efficiency of molecules in solar panels.

In each run, we separate a random 20\% of the datapoints to serve as test data. The experiments were repeated 5 times.
The optimization is carried out using Adam \citep{adam} with the default learning rate (0.01) over 5000 epoch with minibatch size being $\min(N, 10000)$.

Two models are selected for our experiments. A full DGP with $M=200$ and a decoupled DGP with $M_a=500$, $M_b=100$. $M=200$ is a common choice for DGPs and we choose the parameters of the decoupled model to have a proportionate runtime. We use an RBF kernel with a separate lengthscale per dimension in order to have comparable results to \citep{deepgpep} and \citep{deepgpdoubly}. The layer width is fixed at 10 and the depth varies from 0 to 4 hidden layers.

We use the added static mean function per layer as is suggested in \citep{deepgpdoubly}. Adding the layer inputs to the layer outputs as a fixed mean function helps to avoid degenerate covariance matrices and it provides a decent initialization method for the inducing inputs.

\begin{figure}
\begin{minipage}{0.33\textwidth}
\centering
\includegraphics[height=4.4cm]{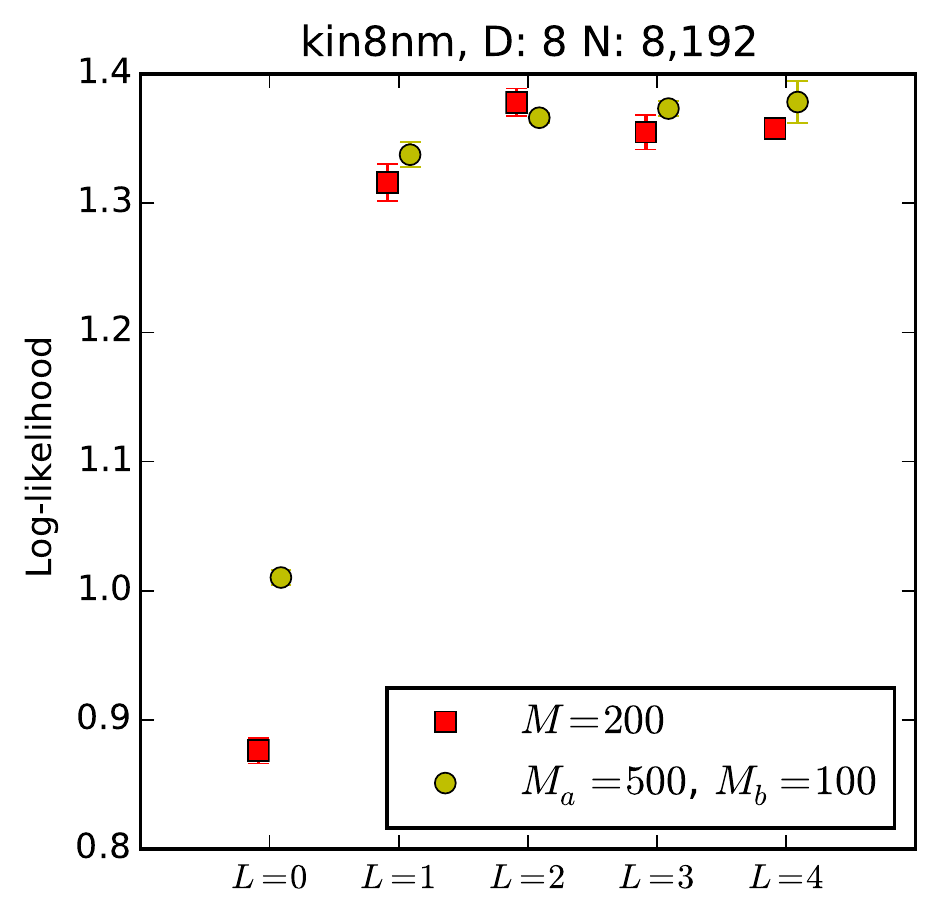}
\end{minipage}
\begin{minipage}{0.33\textwidth}
\centering
\includegraphics[height=4.4cm]{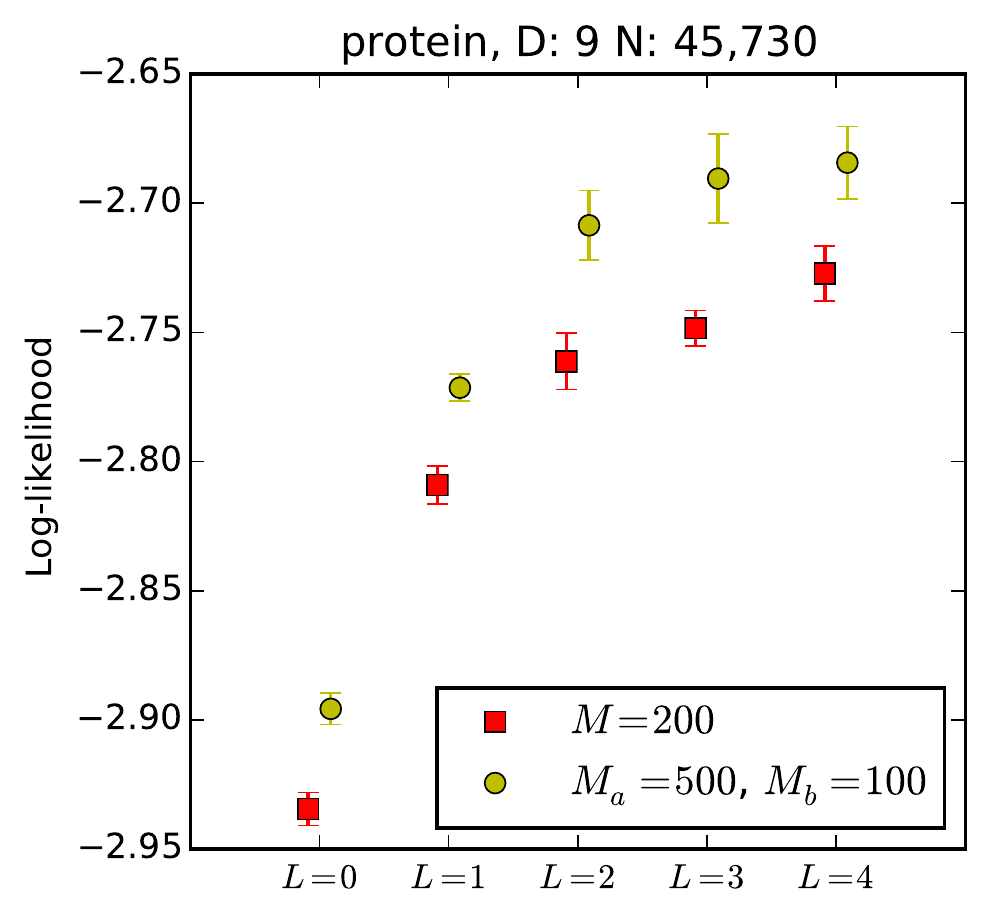}
\end{minipage}
\begin{minipage}{0.33\textwidth}
\centering
\includegraphics[height=4.4cm]{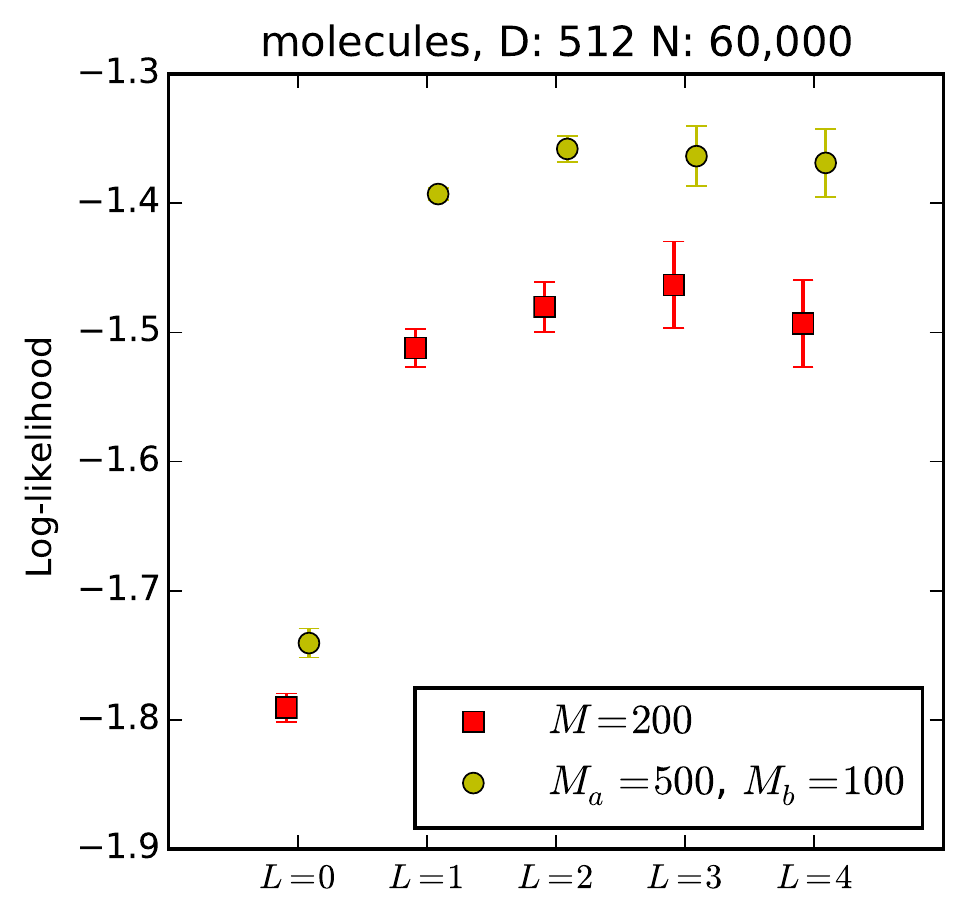}
\end{minipage}
\caption{The test mean log-likelihood of the models. The confidence bands denote one standard deviation. Higher is better.}
\label{ll}
\end{figure}

The results are shown on Figures \ref{ll}, \ref{rmse} with Table \ref{timing} containing the median runtimes. The exact numerical values can be found in Appendix \ref{num}.

\begin{figure}
\begin{minipage}{0.33\textwidth}
\centering
\includegraphics[height=4.4cm]{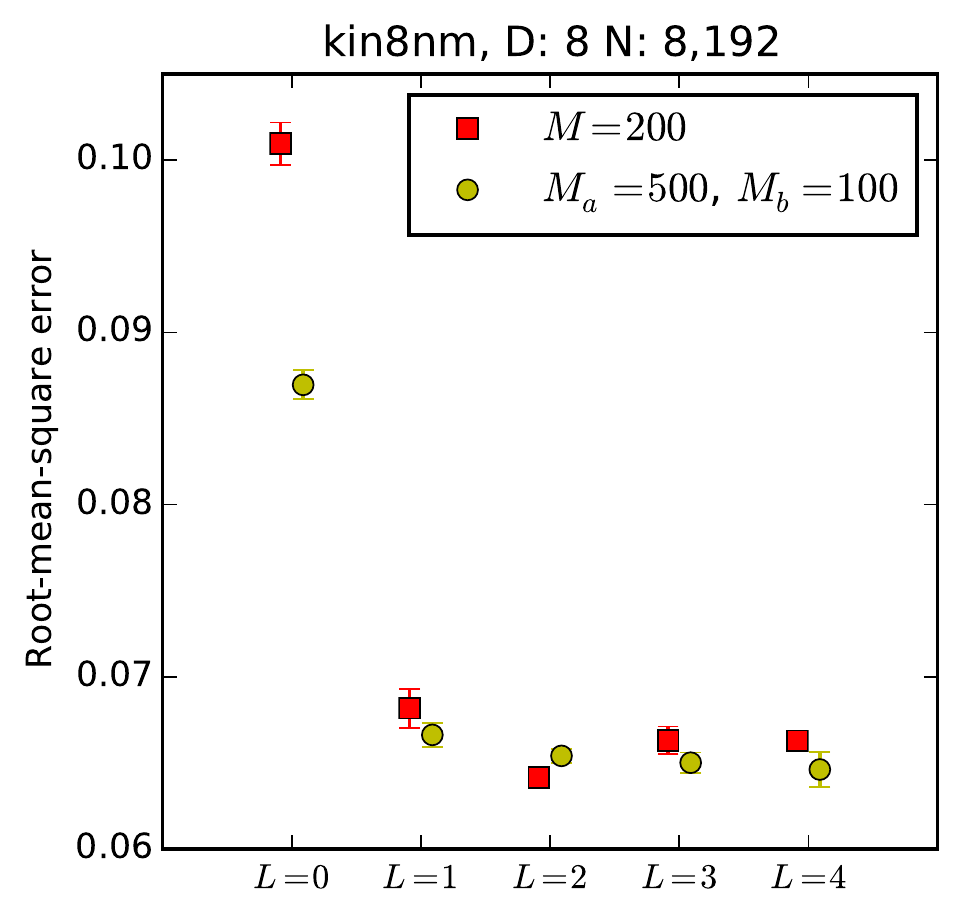}
\end{minipage}
\begin{minipage}{0.33\textwidth}
\centering
\includegraphics[height=4.4cm]{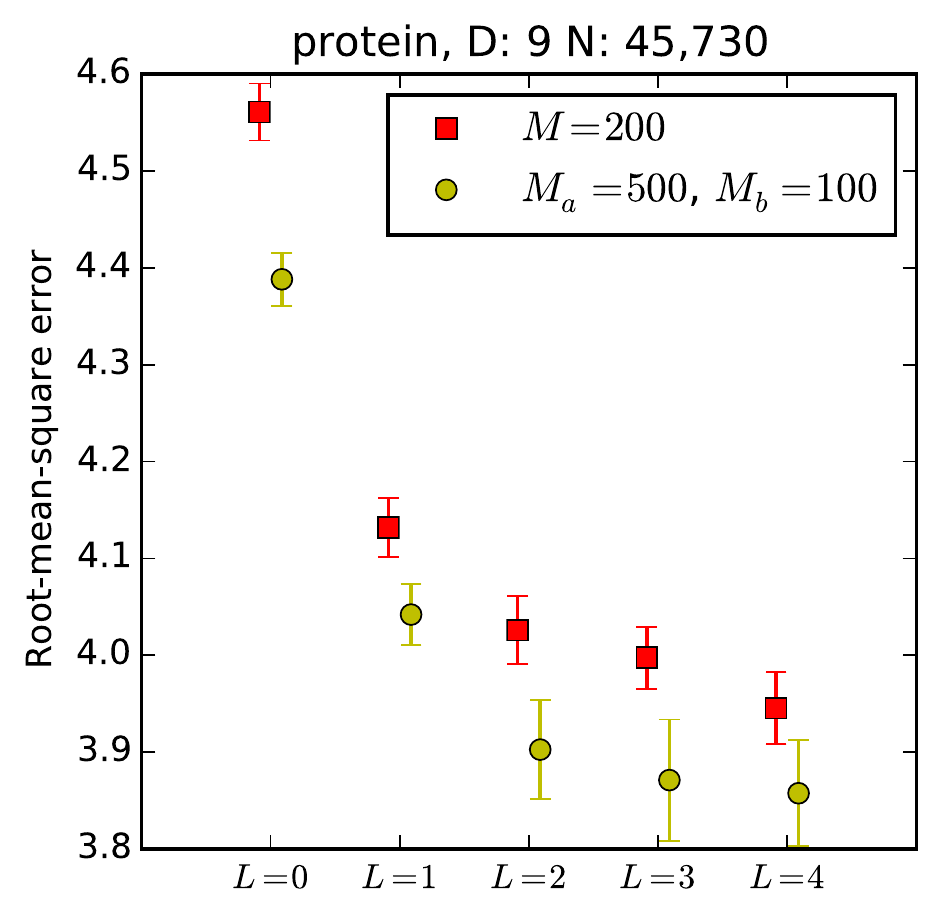}
\end{minipage}
\begin{minipage}{0.33\textwidth}
\centering
\includegraphics[height=4.4cm]{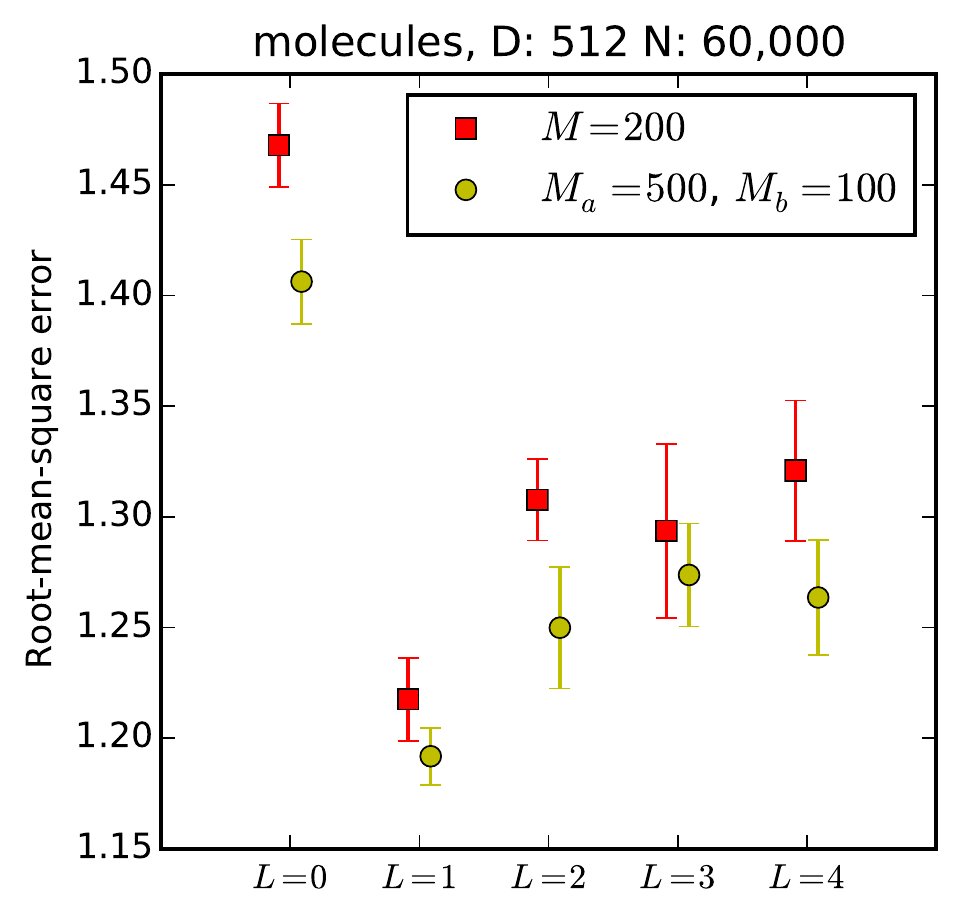}
\end{minipage}
\caption{The test root-mean-square error of the models. The confidence bands denote one standard deviation. Lower is better.}
\label{rmse}
\end{figure}

\begin{table}
\begin{tabular}{lllll}
\toprule
\multicolumn{2}{c}{Model} & kin8nm & protein & molecules \\
\cmidrule(r){1-2} \cmidrule(lr){3-3} \cmidrule(lr){4-4} \cmidrule(l){5-5} 
 $L=0$ & $M=200$              &  \bf 116 s &\bf 587 s &\bf 1699 s \\
       & $M_a=500$, $M_b=100$ & 310 s & 1210 s &2443 s \\
 $L=1$ & $M=200$              &  481 s &2567 s &4159 s \\
       & $M_a=500$, $M_b=100$ &  \bf 433 s &\bf 2015 s &\bf 3683 s \\
 $L=2$ & $M=200$              & 853 s &4577 s &6775 s \\
       & $M_a=500$, $M_b=100$ & \bf 611 s &\bf 3038 s &\bf 5121 s \\
 $L=3$ & $M=200$              & 1242 s &6598 s &9426 s \\
       & $M_a=500$, $M_b=100$ & \bf 811 s & \bf 4111 s &\bf 6596 s \\
 $L=4$ & $M=200$              & 1601 s &8251 s &12122 s \\
       & $M_a=500$, $M_b=100$ & \bf 1017 s &\bf 5209 s & \bf 8078 s \\
       \bottomrule
\end{tabular}
\centering
\caption{Median runtimes. The decoupled model was more cost efficient in the presence of at least one hidden layer.}
\label{timing}
\end{table}

The decoupled DGP outperforms the baseline DGP in every dataset with the exception of the DGP with two hidden layers on kin8nm. Moreover, the runtime of the former is also lower for models with at least one hidden layer. It has higher runtime in the case of the shallow model because in the absence of hidden layers the cost of matrix inversion has a more significant impact on the overall runtime.

\section{Conclusions}

We showed that decoupling the inducing inputs for the mean and the variance is compatible with DGPs and it improves the performance. While we were not able to achieve $O\big(L(DNM_a + DNM_b^2 + M_b^3)\big)$ time complexity due to poor convergence characteristics, we attained a higher, $O\big(L(DNM_a + M_a^3 + DNM_b^2 + M_b^3)\big)$ complexity bound, whose limit is equal to the former for large batch sizes and wide layers.

We exhibited that the decoupled DGPs are both faster and more accurate than the full DGPs. The performance improvement was demonstrated across three distinct datasets. The decoupled model ran faster and achieved a higher log-likelihood and lower root mean squared error.

In future works, we aim to devise a strategy for determining the inducing input locations. One issue that we uncovered is that the inducing inputs suffer from vanishing gradients. We want to experiment with approaches other than stochastic gradient descent to attain inducing inputs that are able to better describe the data.

\section*{Acknowledgment}

The authors would like to acknowledge the generous support from EPSRC.

\clearpage

\bibliographystyle{abbrvnat}
\bibliography{references}

\clearpage

\appendix

\section{Numerical results} \label{num}

Tables \ref{kin8nm_num}, \ref{protein_num}, \ref{molecules_num} contain the numerical results of the experiments. The runtime is the median time of the 5 runs on a Tesla K80 GPU.

\begin{table}[h]
\caption{kin8nm, D: 9 N: 8,192}
\label{kin8nm_num}
\begin{tabular}{llr@{$\pm$}rr@{$\pm$}rr@{ s}}
\toprule
\multicolumn{2}{c}{Model} & \multicolumn{2}{c}{Mean LL} & \multicolumn{2}{c}{RMSE} & \multicolumn{1}{c}{Runtime} \\
\cmidrule(r){1-2} \cmidrule(lr){3-4} \cmidrule(lr){5-6} \cmidrule(l){7-7} 
 $L=0$ & $M=200$              & 0.88 & 0.02 & 0.10 & 0.00 &  \bf 116 \\
       & $M_a=500$, $M_b=100$ & \bf 1.01 & 0.01 & \bf 0.09 & 0.00 &  310 \\
 $L=1$ & $M=200$              & 1.32 & 0.03 & 0.07 & 0.00 &  481 \\
       & $M_a=500$, $M_b=100$ & \bf 1.34 & 0.02 & \bf 0.07 & 0.00 &  \bf 433 \\
 $L=2$ & $M=200$              & \bf 1.38 & 0.02 & \bf 0.06 & 0.00 &  853 \\
       & $M_a=500$, $M_b=100$ & 1.37 & 0.01 & 0.07 & 0.00 &  \bf 611 \\
 $L=3$ & $M=200$              & 1.36 & 0.03 & 0.07 & 0.00 & 1242 \\
       & $M_a=500$, $M_b=100$ & \bf 1.37 & 0.01 & \bf 0.07 & 0.00 &  \bf 811 \\
 $L=4$ & $M=200$              & 1.36 & 0.01 & 0.07 & 0.00 & 1601 \\
       & $M_a=500$, $M_b=100$ & \bf 1.38 & 0.03 &\bf  0.06 & 0.00 & \bf 1017 \\
       \bottomrule
\end{tabular}
\end{table}

\begin{table}[h]
\caption{protein, D: 9 N: 45,730}
\label{protein_num}
\begin{tabular}{llr@{$\pm$}rr@{$\pm$}rr@{ s}}
\toprule
\multicolumn{2}{c}{Model} & \multicolumn{2}{c}{Mean LL} & \multicolumn{2}{c}{RMSE} & \multicolumn{1}{c}{Runtime} \\
\cmidrule(r){1-2} \cmidrule(lr){3-4} \cmidrule(lr){5-6} \cmidrule(l){7-7} 
 $L=0$ & $M=200$              & -2.93 & 0.01 & 4.56 & 0.06 &  \bf 587 \\
       & $M_a=500$, $M_b=100$ & \bf -2.90 & 0.01 & \bf 4.39 & 0.05 & 1210 \\
 $L=1$ & $M=200$              & -2.81 & 0.01 & 4.13 & 0.06 & 2567 \\
       & $M_a=500$, $M_b=100$ & \bf -2.77 & 0.01 & \bf 4.04 & 0.06 & \bf 2015 \\
 $L=2$ & $M=200$              & -2.76 & 0.02 & 4.03 & 0.07 & 4577 \\
       & $M_a=500$, $M_b=100$ & \bf -2.71 & 0.03 & \bf 3.90 & 0.10 & \bf 3038 \\
 $L=3$ & $M=200$              & -2.75 & 0.01 & 4.00 & 0.06 & 6598 \\
       & $M_a=500$, $M_b=100$ & \bf -2.69 & 0.03 & \bf 3.87 & 0.12 & \bf 4111 \\
 $L=4$ & $M=200$              & -2.73 & 0.02 & 3.95 & 0.07 & 8251 \\
       & $M_a=500$, $M_b=100$ & \bf -2.68 & 0.03 & \bf 3.86 & 0.11 & \bf 5209 \\
       \bottomrule
\end{tabular}
\end{table}

\begin{table}[h]
\caption{molecules, D: 512 N: 60,000}
\label{molecules_num}
\begin{tabular}{llr@{$\pm$}rr@{$\pm$}rr@{ s}}
\toprule
\multicolumn{2}{c}{Model} & \multicolumn{2}{c}{Mean LL} & \multicolumn{2}{c}{RMSE} & \multicolumn{1}{c}{Runtime} \\
\cmidrule(r){1-2} \cmidrule(lr){3-4} \cmidrule(lr){5-6} \cmidrule(l){7-7} 
 $L=0$ & $M=200$              & -1.79 & 0.02 & 1.47 & 0.04 &  \bf 1699 \\
       & $M_a=500$, $M_b=100$ & \bf -1.74 & 0.02 & \bf 1.41 & 0.04 &  2443 \\
 $L=1$ & $M=200$              & -1.51 & 0.03 & 1.22 & 0.04 &  4159 \\
       & $M_a=500$, $M_b=100$ & \bf -1.39 & 0.01 & \bf 1.19 & 0.03 &  \bf 3683 \\
 $L=2$ & $M=200$              & -1.48 & 0.04 & 1.31 & 0.04 &  6775 \\
       & $M_a=500$, $M_b=100$ & \bf -1.36 & 0.02 & \bf 1.25 & 0.05 &  \bf 5121 \\
 $L=3$ & $M=200$              & -1.46 & 0.07 & 1.29 & 0.08 &  9426 \\
       & $M_a=500$, $M_b=100$ & \bf -1.36 & 0.05 & \bf 1.27 & 0.05 &  \bf 6596 \\
 $L=4$ & $M=200$              & -1.49 & 0.07 & 1.32 & 0.06 & 12122 \\
       & $M_a=500$, $M_b=100$ & \bf -1.37 & 0.05 & \bf 1.26 & 0.05 &  \bf 8078 \\
       \bottomrule
\end{tabular}
\end{table}

\end{document}